\documentclass[runningheads]{llncs}
\usepackage{graphicx}
\usepackage[backend=biber, maxnames=1, style=lncs, giveninits=true]{biblatex}
\bibliography{refs2.bib}
\usepackage{multirow}
\usepackage[T1]{fontenc}

\begin{document}
\title{Nuclei-Location Based Point Set Registration of Multi-Stained Whole Slide Images}
\titlerunning{Nuclei-Location Based Point Set Registration for WSI}
\author{Adith Jeyasangar\inst{1} \and Abdullah Alsalemi\inst{1} \and
Shan E Ahmed Raza\inst{1} }
\authorrunning{A. Jeyasangar et al.}
\institute{Department of Computer Science, University of Warwick, Coventry, CV4 7AL, UK \\
\email{\{adith.jeyasangar, abdullah.alsalemi, shan.raza\}@warwick.ac.uk}}

\maketitle
\begin{abstract}
Whole Slide Images (WSIs) provide exceptional detail for studying tissue architecture at the cell level. To study tumour microenvironment (TME) with the context of various protein biomarkers and cell sub-types, analysis and registration of features using multi-stained WSIs is often required. Multi-stained WSI pairs normally suffer from rigid and non-rigid deformities in addition to slide artefacts and control tissue which present challenges at precise registration. Traditional registration methods mainly focus on global rigid/non-rigid registration but struggle with aligning slides with complex tissue deformations at the nuclei level. However, nuclei level non-rigid registration is essential for downstream tasks such as cell sub-type analysis in the context of protein biomarker signatures. This paper focuses on local level non-rigid registration using a nuclei-location based point set registration approach for aligning multi-stained WSIs. We exploit the spatial distribution of nuclei that is prominent and consistent (to a large level) across different stains to establish a spatial correspondence. We evaluate our approach using the HYRECO dataset consisting of 54 re-stained images of H\&E and PHH3 image pairs. The approach can be extended to other IHC and IF stained WSIs considering a good nuclei detection algorithm is accessible. The performance of the model is tested against established registration algorithms and is shown to outperform the model for nuclei level registration.
\keywords{Computational pathology \and Nuclei detection \and Point set registration \and Rigid registration \and Non-rigid registration \and Thin plate spline}
\end{abstract}

\section{Introduction}

Digital pathology has revolutionised how pathologists analyse tissue tiles. With the help of digital scanners, biopsies and tissue sections can be scanned digitally as Whole Slide Images (WSIs). These WSIs provide a high resolution digital representation of the entire tissue section allowing pathologists and computational pathology researchers to gain highly-detailed insights into the tissue for more informed cancer diagnosis, prognosis, and treatment \cite{azam_digital_2024}. Moreover, WSIs provide an opportunity to analyse tissue sections with exceptional detail using advanced computational algorithms in computational pathology. 

The most common stain in histology is Haematoxylin and Eosin (H\&E), which highlights nuclei with a blue tint and paints the extracellular matrix with shades of pink \cite{fischer_hematoxylin_2008}. Immunohistochemical (IHC) stains are utilised to stain antibodies to react with specific antigens within a tissue section \cite{ortiz_hidalgo_immunohistochemistry_2022} which can provide deeper insight on the tissue section at hand.

To study spatial relationships among various cell types in the tumour microenvironment with information about the protein biomarker signatures in context, multi-stained WSI pairs are required which aligns H\&E  slide to the corresponding IHC stains at the cell level. These multistained WSI image pairs can be acquired using serial sections or serial staining \cite{lotz_hyreco_2021}.  However, challenges in WSI image registration are numerous, including elastic deformations caused by bleaching and re-staining, tissue folds, torn tissue parts, missing tissue, and other artefacts and deformities. Such obstacles require a registration method that both performs linear and non-linear operations in order to match the moving (source) image with the fixed (target) image. The method should also effectively incorporate information from different stains and types of tissues.

Recent advancements have explored automated image registration techniques to address these limitations. Various researchers have employed intensity-based and feature-based approaches for WSI registration \cite{m_feature_2017}, \cite{wodzinski_multistep_2020}. Feature based registration, primarily focuses on identifying landmarks, and matches these landmarks between the images. However, such techniques often struggle with densely stained regions, where landmarks become obscured. Another major issue is that not all IHC stains produce consistent features as some antibodies stain nuclei, others stain cytoplasmic area and some extracellular structures. Another major limitation of existing approaches is that alignment of the tissue at the local nuclei level is not fully addressed, which hinders studying the spatial relationships between nuclei in multi-stained images.

In this paper, we propose a method for robust alignment of H\&E and IHC stained WSIs incorporating local deformations into account. For this work, we only focus on the Phosphohistone H3 (PHH3) stain from the HYbrid RE-stained and COnsecutive histological serial sections (HYRECO) dataset \cite{lotz_hyreco_2021} which highlights mitotic cells. However, the proposed method can easily be extended to any other IHC markers after training a nuclei detection algorithm.  The method employs a point set registration pipeline for aligning WSIs at the cell level that leverages nuclei as landmarks present in all cells. Nuclei are detected across various tissue types and exhibit minimal stain variations between H\&E and PHH3 tiles. Hence, we can take advantage of such nuclei as a point set from each WSI and accordingly use a point set registration model to solve for the rigid and non-rigid alignment between the points. 
We employ a point set registration method to successfully align PHH3 stained WSI sections with the H\&E stained images followed by a non-rigid B-spline model to further refine the alignment.
The main contributions of this work are summarised as follows:
\begin{enumerate}
    \item We introduce the concept of nuclei-location based point set registration which generalises to H\&E and PHH3 stains; and
    \item We propose a pipeline for WSI image registration that outperforms current established method(s) in local nuclear level alignment comprising multi-step rigid and non-rigid transformations evaluated on the HYRECO dataset.
\end{enumerate}

The remainder of this paper is organised as follows. Section \ref{sec:related-work} briefly outlines related work. Section \ref{sec:methods} delves into the employed dataset and proposed registration pipeline. Section \ref{sec:results} reports and discusses results and benchmarks it against a related method in the literature. Section \ref{sec:conclusions} concludes the paper with a blueprint for future work.

\section{Related Work \label{sec:related-work}}

The two common approaches to image registration are intensity and feature based registration. Traditional methods extract such features, e.g., using Scale-Invariant Feature Transform (SIFT) to estimate the required transformation \cite{hoque_whole_2022}. Recent literature has also explored the notion of using deep learning to extract features from images that aid in formulating the transformation function. A commonly used tool is SuperPoint \cite{DeTone_2018_SuperPoint}, which is a self-supervised CNN-based feature detection model. Also, the former is usually accompanied by SuperGlue \cite{Sarlin_2020_SuperGlue}, an interest point matching model that relies on graph neural networks and optimal matching. 

In this direction, in Awan et al. \cite{awan_deep_2022}, the method proposed a Deep Feature Based Registration (DFBR) feature extractor comprising a pre-processing pipeline that employs foreground tissue segmentation, three layers of a pre-trained VGG-16 deep Convolutional Neural Network (CNN) model, and a non-rigid registration pipeline. The points gathered using this extractor are matched using Euclidean distances between the feature points. These matched points can then be used to calculate an affine transformation between the moving and fixed image. This model has been benchmarked against a similar feature based approaches, producing favourable performance. Therefore, DFBR provides a strong comparative model to compare with our proposed method. Notwithstanding, such model comes with its own drawbacks. A core pre-processing step in DFBR is foreground tissue segmentation, which uses a fine-tuned CNN model, which does not generalise to other datasets without further fine-tuning. Another drawback is that it only focuses on local affine transformation and global non-rigid transformation but does not support local non-rigid transformation to align the nuclei.

Another recent approach is an unsupervised learning model that utilises a spatial transformer, developed by Lee et al. \cite{lee_image-and-spatial_2019}. The proposed image and spatial transformer network implements the notion of Structure of Interest (SoI) in medical images to learn new image representations. By using a combination of a spatial transformer and iterative refinement strategies, this model dynamically adjusts the transformation parameters to gain an accurate alignment. The advantages of this model is the flexibility and adaptability which is well-suited for medical imaging. A limitation for the image and spatial transformer model is that it does not explicitly learn from the landmarks in the images, and therefore, limits generalisability to the unseen images.

Wodzinski et al. \cite{wodzinski_aghsso_nodate} proposed a method relying on SuperPoint and SIFT for keypoint extraction and SuperGlue and RANSAC for feature matching. By iterations, an affine registration is applied and then optimised by non-rigid multi-scale instance optimisation using local normalised cross-correlation. 

Gatenbee et al. \cite{gatenbee2023virtual} developed the Virtual Alignment of PathoLogy Image Series (VALIS) histology registration open-source library. VALIS can perform rigid and non-rigid registration at multiple resolutions and to more than an image pair at a time. The underlying method employs multiple techniques in pre-processing, rigid registration and non-rigid registration to optimise performance, e.g., by combining use of both deep learning models such as VGG and hand-crafted feature descriptors such as Binary Robust Invariant Scalable Keypoints (BRISK) \cite{leutenegger2011brisk}. The main advantage of VALIS is its open-source nature and extensive documentation of the method design and implementation on a plethora of WSI stains and multiple datasets.

On the other hand, Budelmann et al. \cite{budelmann_histokatfusion_2022} presents an optimisation method that relies Newtonian registration. Similar to DFBR, the method segments WSI foregrounds using a CNN. Following automatic rotation alignment through centre of mass is performed. Then affine registration is computed using the Normalised Gradient Fields (NGF) objective function. A deformable registration employs the curvature regularisation function in combination with NGF to estimate local deformities at WSI resolution up to 2 µm/pixel, achieving 155.29 µm of median 90th percentile of TRE.

In the grand scheme of histology image registration, a nuclei-based point set approach to image registration aims to address some of the challenges seen in existing methods. Unlike some of the discussed literature, the proposed registration model does not require large datasets for training, with exception to the nuclei detection model that requires large number of nuclei-labelled images. However, we can use various publicly available pre-trained deep learning models for nuclei detection such as Hover-Net \cite{graham_hover-net_2019}. Furthermore, our point set registration model can be more generalisable than most neural network based approaches as it relies on detection nuclei for estimation of the transformation function. Also, our model includes both local rigid and non-rigid transformations, leading to a higher precision in alignment. The most important factor is the downstream analysis, as most of the downstream analysis is done at the nuclei level, the registration should be accurate to the nuclei level to collect nuclei-level features from multiple stains.

\section{Methods}
\label{sec:methods}

\subsection{Dataset}
The HYRECO dataset \cite{lotz_hyreco_2021} is a publicly available WSI registration dataset acquired at the Radboud University Medical Center containing two data subsets. Subset A compromises of nine different cases with four consecutive slide using four different stains: H\&E, CD8, KI67, CD45. Subset B is made up of 54 re-stained image pairs of H\&E and PHH3. In addition, PHH3 stains are achieved by removing the cover slip from the H\&E stain, bleaching the tissue and re-staining the same tissue slice with PHH3. There are 2,303 point annotations across the 54 images pairs, resulting in about 43 annotations per image pair.

In evaluating the proposed pipeline, Subset B is used, focusing on the H\&E and PHH3 stains. The significance of choosing such stains is that all nuclei are clearly identifiable which helps with initial validation of our approach. However, it is important to mention that this approach can easily be transferred to other stains, once the overall pipeline is optimised for one stain. 

\begin{figure}
\centering
\makebox[\textwidth][c]{\includegraphics[width=120mm]{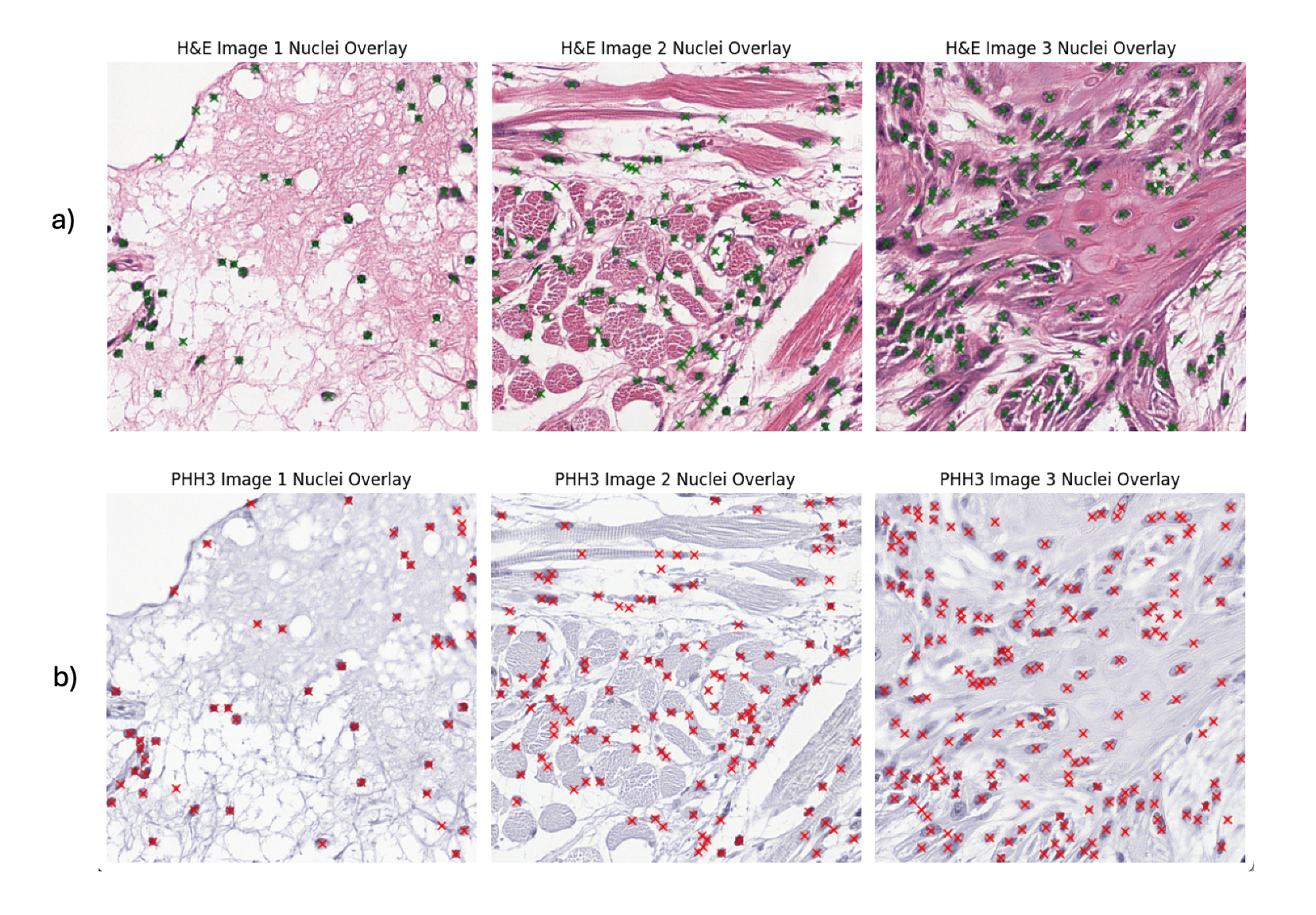}}
\caption{Tiles extracted from H\&E and PHH3 slides. Row a) Shows H\&E slides with green cross indicating nuclei. Row b) Shows PHH3 slides with red cross indicating nuclei.\label{nuclei_overlay}}
\end{figure}

Although there is a slight variation in the shade of the nuclei between two images, there is a clear distinction of nuclei in both images, allowing the nuclei detection algorithm to detect most nuclei in the image. Accordingly, we have employed a pre-trained Hover-Net from the TIAToolbox \cite{pocock2022tiatoolbox}, which is a widely-known nuclei detection, segmentation, and classification model trained on H\&E images \cite{graham_hover-net_2019}. In order to test its efficacy on PHH3 images, a manual check was carried out to verify whether a given point can be detected in both corresponding H\&E and PHH3 tiles, as shown in Fig.~\ref{nuclei_overlay}.

\subsection{Evaluation Metrics}
To measure the performance of the registration model, we use a widely adopted metric is Target Registration Error (TRE) as suggested in the HYRECO dataset paper \cite{lotz_comparison_2023}. TRE utilises landmarks in both the fixed and moving images to compute the distance error between them. This can be calculated using (\ref{equ:tre}):

\begin{equation}\label{equ:tre}
TRE = ||t_k - f_k||_2
\end{equation}

where $t_k, f_k$ are the available landmarks in both images. We can also normalise the TRE value using the diagonal shown in (\ref{equ:rtre}):
\begin{equation}\label{equ:rtre}
rTRE = TRE/ (\sqrt{w^2 +h ^2})
\end{equation}
where $w$ and $h$ are the width and height of the fixed image respectively.

Commonly, the landmarks specified in the HYRECO dataset encompasses all the images that can be used to calculate TRE. However, in this work, as we focus on local non-rigid registration our specified size of image tiles is $1024$ $\times1024$ pixel. Therefore, it is likely that there would only a few landmarks in each tile from the original dataset. Therefore, evaluating the model using error distance between a few landmarks is not a robust evaluation method. 

We adopt a more effective approach to TRE computation, by using nuclei detections instead of landmarks on the warped image (i.e., the image created from transformation of moving image) as this gives a high number of landmark points. We can then compute TRE to evaluate the distance error between the warped image nuclei and the fixed image's in terms of the detected nuclei positions.

We also calculate the Average rTRE (ArTRE) and Median rTRE (MrTRE) to provide a more comprehensive understanding of the performance. ArTRE gives the mean accuracy value of the typical registration error, which is more influenced by extreme values. Though, MrTRE is not influenced by such extreme values. Given both these values, we have a comprehensive evaluation procedure to estimate registration performances across all image pairs.

\subsection{Overview of Registration Pipeline}
\begin{figure}
\makebox[\textwidth][c]{\includegraphics[width=120mm]{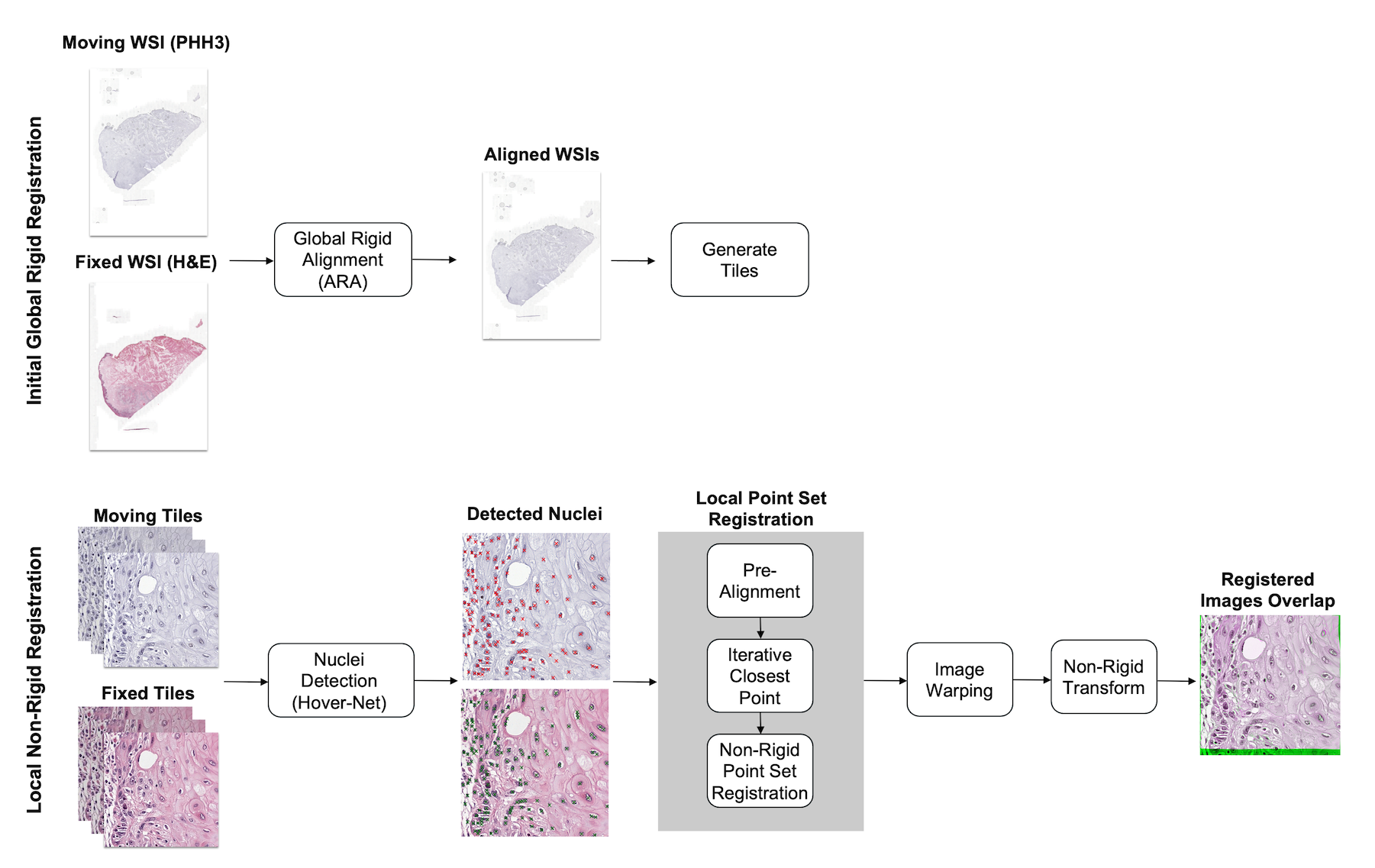}}
\centering
\caption{Overview of nuclei-location based point set registration pipeline. First, the WSI pair is roughly aligned by a rigid registration step. Second, image tiles are extracted from both images where a fixed and moving tile undergoes a pre-processing stage followed by nuclei detection. Following, the point-set registration model carries out pre-alignment, ICP, and non-rigid point-set registration operations. Once a transformation matrix is obtained, it is applied to the moving image, producing a wrapped image. A further refined non-rigid registration step using the B-Spline method is conducted to obtained the final result. \label{pipeline}}
\end{figure}

We propose a four stage pipeline to accurately align WSIs as extracted tiles of $1024$ $\times1024$ pixels resolution. This pipeline consists of global rigid registration followed by local nuclei detection, point set registration, image warping and non-rigid registration, as illustrated in Fig.~\ref{pipeline}. 
\newpage
During the nuclei detection stage, we generate a binary mask that represents the position of nuclei so that the point set can be extracted in preparation for the point set registration model. The point set registration model carries out the following operations: Automatic Rotation Alignment (ARA), optimisation using Iterative Closest Point (ICP) and non-rigid point set registration. After the point set registration model, the parameters gathered are then applied to the moving image to obtain a warped image followed by a further B-spline non-linear registration method for further improvement in alignment performance. \\

As previously mentioned, we focus primarily on local nuclei level non-rigid registration. For local registration to work, the WSI needs to be pre-aligned. We  employ Automatic Rotation Alignment described in Deep feature based cross slide registration \cite{awan_deep_2022}, \cite{lotz_robust_2019} for pre-alignment. This provides a rough alignment between WSIs to extract tiles from the same location in the WSI pair. This allows for similar nuclei landmarks to be equally detected in both tiles so that the nuclei-location based point set registration model can focus on local registration. The primary focus of this model is to work with H\&E and PHH3 stained WSIs, however, the pipeline can be adapted to work with other IHC stains. 

\subsubsection{Nuclei Detection}
Nuclei detection is a foundational step in our proposed method. Accurate detection of nuclei is essential as it serves as a point set used in the point set registration model. In this work, we employ a pre-trained Hover-Net model from the TIAToolbox \cite{pocock2022tiatoolbox}. The Hover-Net architecture \cite{graham_hover-net_2019} is well-suited for simultaneous segmentation and classification of nuclei in WSIs using a CNN to detect, segment, and identify types of nuclei that make up a WSI tile. However, more optimised nuclei detection models such as SCCNN and MapDe can also be utilised \cite{sirinukunwattana2016locality, raza2019deconvolving}.

Using Hover-Net for nuclei detection demonstrated superior performance compared to other pre-existing detection algorithms \cite{naylor_segmentation_2019} for H\&E stained images. The improved accuracy translates to a more robust registration process as the point set provided by the nuclei detection model is less prone to noise and anomalous detections. Accordingly, the model outputs a binary mask, where an active pixel represents a nucleus location and inactive pixel represents otherwise. We can then use this binary mask to create a point set for the registration model by finding the coordinates in the binary mask where the corresponding pixel is active. This will allow two point sets to be created: (a) the fixed point set comprising of the nuclei points in the H\&E stains and (b) the moving point set comprising of the nuclei points in the PHH3 stains images. 

\subsection{Rigid Registration}

After retrieving the two point sets during nuclei detection, they then undergo a three-step operation for point set registration to gather the required transformation parameters for slide tile warping.

\subsubsection{Automatic Rotation Alignment (ARA)}
Differently-stained tissues can have different positions and rotations in the scanned WSI. Therefore, an initial translation and rotation of the point set is performed to correct the rigid alignment between the fixed and moving point sets.

ARA \cite{awan_deep_2022},\cite{lotz_robust_2019} is computed by determining the translation between the fixed and moving point set. This model uses two methods used to calculate rough translation: The first one is using the centre of mass of the two point sets to find the suitable translation vector. This is the most common approach described in the literature, however, it may lead to inconsistent results, especially in terms of registration point sets. Therefore, phase correlation is applied simultaneously with centre of mass translation to find the most accurate translation with the lowest Mean Square Error (MSE) between the points. The translation vector is then used to translate the moving point set. MSE is computed in (\ref{equ:mse}):

\begin{equation}
\label{equ:mse}
MSE = \frac{1}{n} \Sigma (Y_i - \hat{Y_i})^2
\end{equation}

where $(Y_i - \hat{Y_i})^2$ is the square of the error between a matched point in point sets.

The next step is a rotation operation to help find the global minimum alignment. Angle values equidistantly sampled from $[0, 360]$ degrees are used to rotate the moving point set. For all point sets created, the MSE metric is calculated and used to find the angle with the lowest MSE value. This provides a rough estimate between the two point sets. 

\subsubsection{Iterative Closest Point (ICP)}
At each rotation step in the ARA, the ICP algorithm is employed to further refine the precision of the translation and rotation. ICP iteratively adjusts the affine transformation between matched points to achieve a more precise rigid transformation. The ICP algorithm is described as follows:

\begin{itemize}
    \item Pairs of points are selected from the moving and fixed point sets whose Euclidean distance falls below a specified distance threshold. KD-Tree \cite{bentley1975multidimensional} is used for efficient search of closest point pairs. 
    \item Use pairs of point to compute a rigid transformation between points and creates a rotation matrix and translation vector.
    \item The rotation matrix and translation vector is applied to the moving point set for refined alignment of point sets.
    \item Iteratively align the point set until the maximum number of iterations is reached or the distance threshold is met.
\end{itemize}

The disadvantage of the ICP algorithm is that it converges to a local minimum, however, it is important that we find the global minimum for accurate points set registration. Combing ICP with ARA, ensures that such combination results in finding the global minimum. 

\subsection{Non-Rigid Registration}

\subsubsection{Point Set Registration}
The final stage of the point set registration model focuses on aligning local deformations in the nuclei point set for non-rigid registration. 

In this step, a simplified Gaussian Mixture Model based point set registration is presented in \cite{zhang_point_2017}, \cite{ge_non-rigid_2014}. The method deals with non-rigid transformations in point sets. Thus, we create a Locally Linear Embedding  (LLE) weight matrix which contains the neighbours of each point in the moving image point set. LLE algorithm proposes a non-linear dimensionality reduction method to preserve the local neighborhood structure in the lower dimensional space. This is carried out by computing the K-nearest neighbours of each point in the moving point set based on Euclidean distances, resulting in a LLE matrix that represents each point in the moving point set as a linear combination of its neighbours. Then it computes the Gaussian kernel matrix used in Coherent Point Drift (CPD) algorithm to estimate the non-rigid transformation for each point. This can then be used to solve a linear system to obtain a weight matrix $W$. Finally, the matrix is used to transform the moving point set. This step continues iterating until the convergence criteria is met.

\subsubsection{Image Warping}
After completing registration on the moving image point set, we need to apply both linear and non-linear transformations onto the image tile to obtain the registered image. In this stage, we can use a simple transformation matrix created using the rotation angle and translation vector gathered in the ARA and ICP algorithms. This transformation matrix can be applied onto the image to obtain the rigid alignment. To apply the non-rigid alignment, the Thin Plate Spline (TPS) technique provides a robust method in applying the non-rigid transformation found between the point sets, creating a deformation field. It is important to note that the non-rigid transformation considers the local deformations present in the images caused by the tissue variations. Control points can be assigned to the point set before the non-rigid registration model and the warped control points be the matched points after the registration. The smoothness of the deformation field created ensures that the nuclei in the moving image is mapped correctly to the nuclei in the fixed image. 

\subsubsection{B-Spline Non-Rigid Registration}
\begin{figure}
\makebox[\textwidth][c]{\includegraphics[width=0.6\textwidth]{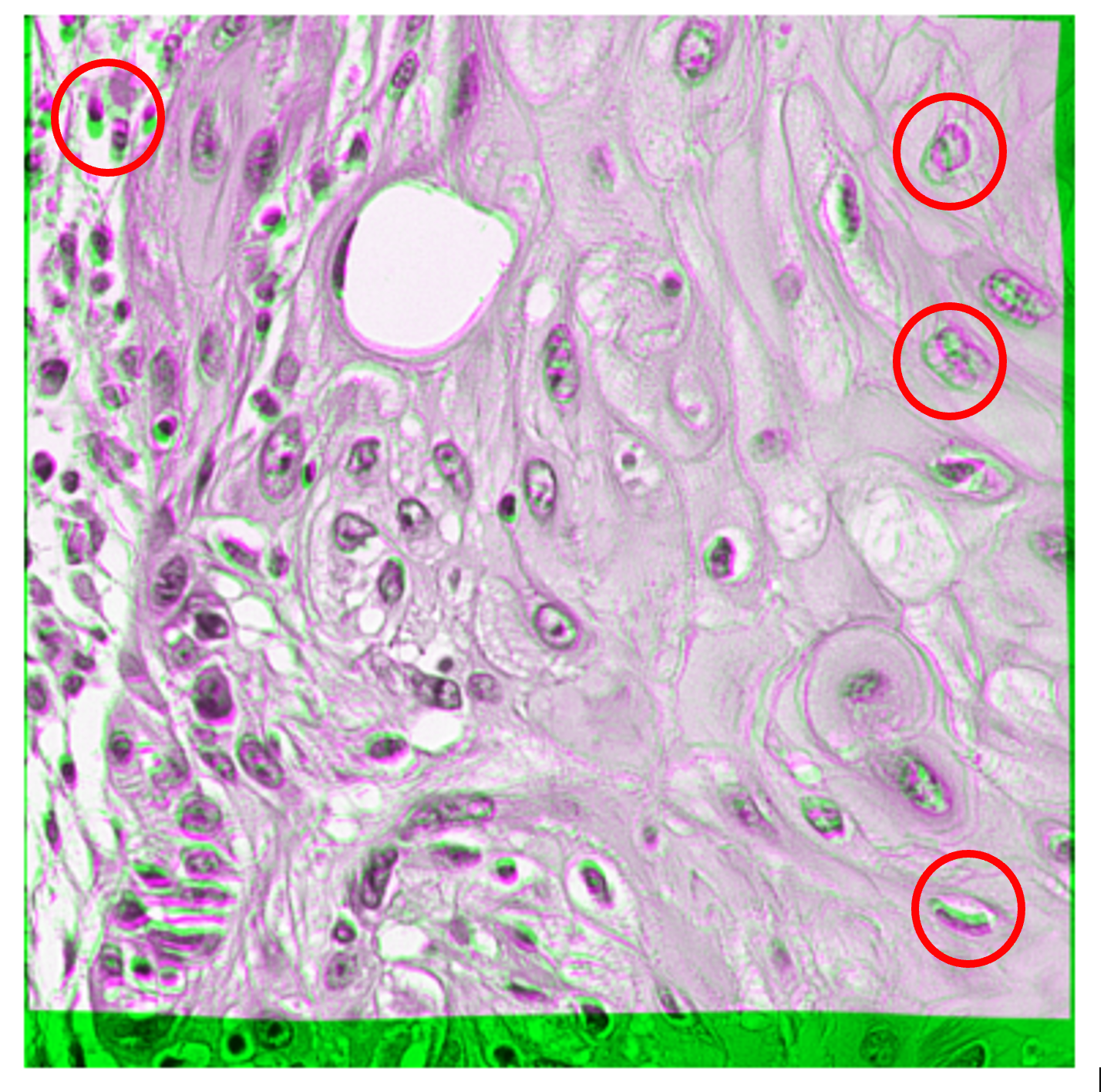}}%
\centering
\caption{Overlay of a sample fixed and warped image tiles using the proposed pipeline. The red circles identifies the mis-alignments that can be fixed by using B-Spline non rigid registration.}\label{fig:over_wo_b-spline}
\end{figure}

To further refine, the wrapped image and reduce registration error, a final stage in the registration pipeline includes b-spline based alignment at the local level.
As shown in the overlay of the warped and fixed image in Fig.~\ref{fig:over_wo_b-spline}, a few nuclei has a small mis-alignment that needs to be amended. This is due to TPS over-fitting, and as a result, not always being able to fully match the control points with the warped control points.

Therefore, a further non-rigid registration operation is applied. The approach chosen employs a multi-resolution B-spline model using SimpleITK library \cite{lowekamp2013design} which uses B-spline curves to define a continuous deformation field that maps data point in moving image to a corresponding data point in the fixed image. Then the deformation field is applied to the images to obtain the final resultant warped image, resulting in a more precisely aligned image tile.

\section{Results and Discussion}
\label{sec:results}

Our proposed method has been evaluated against the publicly available Awan et al.'s DFBR \cite{awan_deep_2022} using two different stained subsets of the HYRECO dataset: H\&E and PHH3. To evaluate the performance of our pipeline, we conduct a series of tests: (a) Evaluate the performance of the nuclei segmentation Hover-Net model; (b) evaluate the accuracy of the point set registration, by testing the model on 100 different $1024$ pixels $\times1024$ pixels  tiles; (c) benchmark the results with DFBR; and (d) show the relationship between the accuracy of the model and the number of nuclei annotated on a given tile.

As shown in Fig.~\ref{nuclei_overlay}, we have performed Hover-Net based nuclei detection on 100 tiles pairs of H\&E and PHH3 stains and manually verifying the accuracy of the alignment. For the majority of the cases, most of the nuclei are detected. However, there were some instances that some nuclei are not detected in PHH3 stained tiles. We observe that the low number of undetected nuclei is relatively insignificant to the overall performance of the point set registration. Also, our pipeline should be robust to these small number of missing nuclei to make it less reliant and more robust to imperfections in nuclei detection algorithm.

Benchmarking the performance of the pipeline is crucial to evaluate how the model performs against state-of-the-art methods in literature. For example, as shown in Table~\ref{tab1}, we compare the proposed model with the DFBR implementation in TIAToolbox including the B-spline non-rigid registration. Both models have been run on 100 H\&E and PHH3 extracted image tile pairs from the HYRECO dataset.

\begin{table}[htb]
\caption{Proposed method results benchmarked against DFBR.}
\centering
\makebox[\textwidth][c]{
\resizebox{1.10\linewidth}{!}{
\label{tab1}
\begin{tabular}{|c|c|c|c|c|}
\hline
\multirow{2}{*}{\textbf{Metric/Method}} & \multicolumn{2}{c|}{\textbf{Average rTRE}} & \multicolumn{2}{c|}{\textbf{Median rTRE}} \\
\cline{2-5}
& Average & Median & Average & Median\\
\hline
\textbf{Point Set+B-Spline} & $\mathbf{\{12.8\pm9.64\}\times10^{-3}}$ & $\mathbf{\{5.91\pm9.64\}\times10^{-3}}$ & $\mathbf{\{7.66\pm6.69\}\times10^{-3}}$ & $\mathbf{\{1.54\pm6.69\}\times10^{-3}}$ \\
\hline
\textbf{Point Set}& $\{16.3\pm12.2\}\times10^{-3}$ & $\{10.4\pm12.2\}\times10^{-3}$ & $\{11.8\pm9.3\}\times10^{-3}$ & $\{6.32\pm9.27\}\times10^{-3}$ \\
\hline
\textbf{DFBR} & $\{22.0\pm7.32\}\times10^{-3}$ & $\{17.6\pm7.32\}\times10^{-3}$ & $\{18.1\pm2.82\}\times10^{-3}$ & $\{14.9\pm2.82\}\times10^{-3}$ \\
\hline
\end{tabular}}
}
\end{table}

\begin{figure}
\makebox[\textwidth][c]{\includegraphics[width=1.0\textwidth]{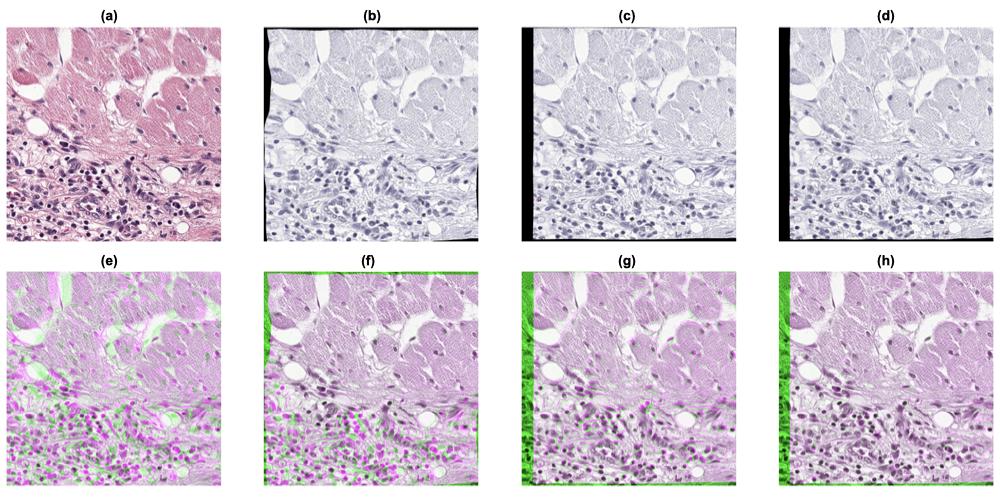}}%
\centering
\caption{Sample tile of a \textbf{(a)} fixed Image, \textbf{(b)} warped image using DFBR, \textbf{(c)} warped image from proposed method, and \textbf{(d)} warped image using proposed method + B-spline, followed by an overlay of fixed image and warped images \textbf{(e)} with original moving image, \textbf{(f)} with warped image using DFBR, \textbf{(g)} with warped image using the proposed method, and \textbf{(h)} with warped image using the proposed method + B-spline.\label{wsi2}}
\end{figure}

Table~\ref{tab1} reports the proposed methods results alongside DFBR. The proposed point set achieves an average rTRE of $\{1.627 \pm 1.224\}\times10^{-2}$ outperforming the DFBR model. In addition, the point set model with B-spline further improves accuracy to an average rTRE of $\{1.283 \pm 0.96\}\times10^{-2}$. Both models outperform DFBR. However, the standard deviation of the proposed model is slightly higher than those of the DFBR model. This shows that there is a higher variability in the accuracy of the proposed methods compared to DFBR, and hence the average rTRE of $2.202\times10^{-2}$ is more consistent between all tested tile images.
This is similar with the Median rTRE. The results are illustrated in Fig. \ref{wsi2}. It is worth noting that we have tested another related method \cite{trahearn2017registration} to benchmark the proposed model further, however, the method did not produce reasonable results due to it being optimised for a specific dataset, among other technical limitations. 

\begin{figure}
\centering
\makebox[\textwidth][c]{\includegraphics[width=0.7\textwidth]{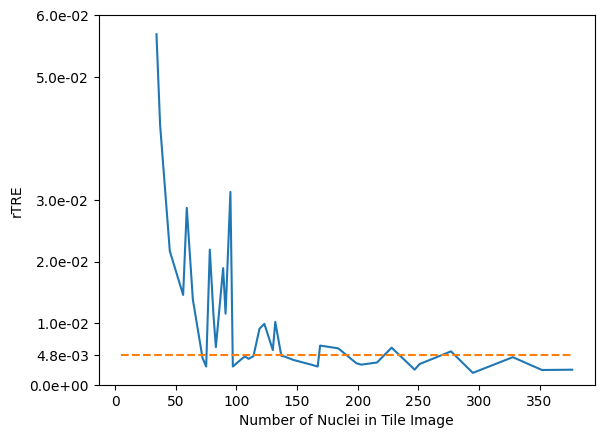}}
\caption{Effect of detected nuclei count on the $rTRE$ of registration pipeline. It is evident that after exceeding 100 nuclei detections, $rTRE$ plateaus around $5\times10^{-3}$.\label{fig4}}
\end{figure}

As the point set registration method heavily relies on the nuclei point set, it is likely that the accuracy of the model is affected by the number of nuclei in the tile. This correlation has been shown in Fig.~\ref{fig4} such that as the nuclei number increases, rTRE decreases. After about 100 nuclei or more in a given image tile, the rTRE value starts to plateau. Therefore, we recommend to choose a tile size which can encompass at least 100 nuclei in the tile for accurate nuclei-based point set registration model. 
\newpage
The computational time of the registration model is also an important factor to evaluate. We tested the models on a machine with Intel Core i5-8500 (6 cores), 64GB RAM and Nvidia RTX 2080Ti GPU (11GB). When run on a $1024$ $\times1024$ pixels image, the first stage of nuclei detection takes around eight seconds to extract the point set from the image pair. The point set registration model takes about six seconds to register and apply to the image. Therefore, it takes about only 14 seconds in total to run the point set registration model per image pair. If the B-spline operation is further applied, it takes an extra 16 seconds leading to a total of 30 seconds per image pair. As compared to the DFBR model, when run it takes 36 seconds running on the same machine. Therefore, point set model slightly improves in speed compared to the DFBR if we improve the nuclei detection and replace Hover-Net with a more efficient nuclei detection algorithm.

\section{Conclusion}
\label{sec:conclusions}

In this paper, we present a novel nuclei-based point set registration pipeline for  histology images. The proposed method can extract the nuclei point set from the WSI tiles using the Hover-Net nuclei detection and segmentation model. The point set registration then computes both rigid and non-rigid transformations between the two obtained point sets. These transformations are then iteratively applied to the image pair through affine transformation matrix and TPS, eventually obtaining a final registered image. Our proposed method achieves a noticeable boost in performance of for local non-rigid registration required for the downstream tasks. We show that our model outperforms an existing model such as the DFBR that utilises deep learning for finding the registration parameters. 

There are a couple of limitations of this approach: 1. it requires a good nuclei detection algorithm as a pre-processing step; and 2. it requires at least 100 nuclei detected per tile. In future, we plan to utilise a computationally efficient and robust pipeline for nuclei detection to improve performance. In addition, we plan to extend this pipeline to perform registration on other IHC stains.

\printbibliography

@article{ortiz_hidalgo_immunohistochemistry_2022,
	title = {Immunohistochemistry in {Historical} {Perspective}: {Knowing} the {Past} to {Understand} the {Present}},
	volume = {2422},
	issn = {1940-6029},
	language = {eng},
	journal = {Methods in Molecular Biology (Clifton, N.J.)},
	author = {Ortiz Hidalgo, Carlos},
	year = {2022},
	pmid = {34859396},
	pages = {17--31},
}

@article{fischer_hematoxylin_2008,
	title = {Hematoxylin and {Eosin} {Staining} of {Tissue} and {Cell} {Sections}},
	volume = {2008},
	issn = {1940-3402, 1559-6095},
	doi = {10.1101/pdb.prot4986},
	number = {5},
	journal = {Cold Spring Harbor Protocols},
	author = {Fischer, Andrew H. and Jacobson, Kenneth A. and Rose, Jack and Zeller, Rolf},
	month = may,
	year = {2008},
	pmid = {21356829},
	pages = {pdb.prot4986},
}

@article{m_feature_2017,
	title = {Feature and {Intensity} {Based} {Medical} {Image} {Registration} {Using} {Particle} {Swarm} {Optimization}},
	volume = {41},
	issn = {1573-689X},
	doi = {10.1007/s10916-017-0846-9},
	number = {12},
	journal = {Journal of medical systems},
	author = {M, Abdel-Basset and Ae, Fakhry and I, El-Henawy and T, Qiu and Ak, Sangaiah},
	month = nov,
	year = {2017},
	pmid = {29098445},
}

@article{wodzinski_multistep_2020,
	title = {Multistep, automatic and nonrigid image registration method for histology samples acquired using multiple stains},
	copyright = {https://iopscience.iop.org/page/copyright},
	issn = {0031-9155, 1361-6560},
	doi = {10.1088/1361-6560/abcad7},
	journal = {Physics in Medicine \& Biology},
	author = {Wodzinski, Marek and Skalski, Andrzej},
	month = nov,
	year = {2020},
}

@article{azam_digital_2024,
	title = {Digital pathology for reporting histopathology samples, including cancer screening samples – definitive evidence from a multisite study},
	volume = {84},
	issn = {1365-2559},
	doi = {10.1111/his.15129},
	language = {en},
	number = {5},
	journal = {Histopathology},
	author = {Azam, Ayesha S and Tsang, Yee-Wah and Thirlwall, Jenny and Kimani, Peter K and Sah, Shatrughan and Gopalakrishnan, Kishore and Boyd, Clinton and Loughrey, Maurice B and Kelly, Paul J and Boyle, David P and Salto-Tellez, Manuel and Clark, David and Ellis, Ian O and Ilyas, Mohammad and Rakha, Emad and Bickers, Adam and Roberts, Ian S D and Soares, Maria F and Neil, Desley A H and Takyi, Abi and Raveendran, Sinthuri and Hero, Emily and Evans, Harriet and Osman, Rania and Fatima, Khunsha and Hughes, Rhian W and McIntosh, Stuart A and Moran, Gordon W and Ortiz-Fernandez-Sordo, Jacobo and Rajpoot, Nasir M and Storey, Ben and Ahmed, Imtiaz and Dunn, Janet A and Hiller, Louise and Snead, David R J},
	year = {2024},
	pages = {847--862},
}

@article{lotz_comparison_2023,
  title={Comparison of consecutive and restained sections for image registration in histopathology},
  author={Lotz, Johannes and Weiss, Nick and van der Laak, Jeroen and Heldmann, Stefan},
  journal={Journal of Medical Imaging},
  volume={10},
  number={6},
  pages={067501--067501},
  year={2023},
  publisher={Society of Photo-Optical Instrumentation Engineers}
}

@article{naylor_segmentation_2019,
  author={Naylor, Peter and Laé, Marick and Reyal, Fabien and Walter, Thomas},
  journal={IEEE Transactions on Medical Imaging}, 
  title={Segmentation of Nuclei in Histopathology Images by Deep Regression of the Distance Map}, 
  year={2019},
  volume={38},
  number={2},
  pages={448-459},
  doi={10.1109/TMI.2018.2865709}}

@misc{lotz_hyreco_2021,
    doi = {10.21227/pzj5-bs61},
    url = {https://dx.doi.org/10.21227/pzj5-bs61},
    author = {van der Laak, Jeroen and Lotz, Johannes and Weiss, Nick and Heldmann, Stefan},
    publisher = {IEEE Dataport},
    title = {HyReCo - Hybrid re-stained and consecutive histological serial sections},
    year = {2021}
}

@inproceedings{ge_non-rigid_2014,
	title = {Non-rigid {Point} {Set} {Registration} with {Global}-{Local} {Topology} {Preservation}},
	doi = {10.1109/CVPRW.2014.45},
	booktitle = {2014 {IEEE} {Conference} on {Computer} {Vision} and {Pattern} {Recognition} {Workshops}},
	author = {Ge, Song and Fan, Guoliang and Ding, Meng},
	month = jun,
	year = {2014},
	pages = {245--251},
}

@inproceedings{zhang_point_2017,
	title = {Point {Set} {Registration} with {Global}-{Local} {Correspondence} and {Transformation} {Estimation}},
	doi = {10.1109/ICCV.2017.291},
	booktitle = {2017 {IEEE} {International} {Conference} on {Computer} {Vision} ({ICCV})},
	author = {Zhang, Su and Yang, Yang and Yang, Kun and Luo, Yi and Ong, Sim Heng},
	month = oct,
	year = {2017},
	pages = {2688--2696},
}

@misc{lotz_robust_2019,
	title = {Robust, fast and accurate: a 3-step method for automatic histological image registration},
	doi = {10.48550/arXiv.1903.12063},
	publisher = {arXiv},
	author = {Lotz, Johannes and Weiss, Nick and Heldmann, Stefan},
	month = mar,
	year = {2019},
}

@misc{graham_hover-net_2019,
  title={Hover-net: Simultaneous segmentation and classification of nuclei in multi-tissue histology images},
  author={Graham, Simon and Vu, Quoc Dang and Raza, Shan E Ahmed and Azam, Ayesha and Tsang, Yee Wah and Kwak, Jin Tae and Rajpoot, Nasir},
  journal={Medical image analysis},
  volume={58},
  pages={101563},
  year={2019},
  publisher={Elsevier}
}

@article{hoque_whole_2022,
	title = {Whole slide image registration via multi-stained feature matching},
	volume = {144},
	issn = {0010-4825},
	doi = {10.1016/j.compbiomed.2022.105301},
	journal = {Computers in Biology and Medicine},
	author = {Hoque, Md. Ziaul and Keskinarkaus, Anja and Nyberg, Pia and Mattila, Taneli and Seppänen, Tapio},
	month = may,
	year = {2022},
	pages = {105301},
}

@misc{lee_image-and-spatial_2019,
  title={Image-and-spatial transformer networks for structure-guided image registration},
  author={Lee, Matthew CH and Oktay, Ozan and Schuh, Andreas and Schaap, Michiel and Glocker, Ben},
  booktitle={Medical Image Computing and Computer Assisted Intervention--MICCAI 2019: 22nd International Conference, Shenzhen, China, October 13--17, 2019, Proceedings, Part II 22},
  pages={337--345},
  year={2019},
  organization={Springer}
}

@misc{awan_deep_2022,
  title={Deep feature based cross-slide registration},
  author={Awan, Ruqayya and Raza, Shan E Ahmed and Lotz, Johannes and Weiss, Nick and Rajpoot, Nasir},
  journal={Computerized Medical Imaging and Graphics},
  volume={104},
  pages={102162},
  year={2023},
  publisher={Elsevier}
}

@article{pocock2022tiatoolbox,
  title={TIAToolbox as an end-to-end library for advanced tissue image analytics},
  author={Pocock, Johnathan and Graham, Simon and Vu, Quoc Dang and Jahanifar, Mostafa and Deshpande, Srijay and Hadjigeorghiou, Giorgos and Shephard, Adam and Bashir, Raja Muhammad Saad and Bilal, Mohsin and Lu, Wenqi and others},
  journal={Communications medicine},
  volume={2},
  number={1},
  pages={120},
  year={2022},
  publisher={Nature Publishing Group UK London}
}

@article{lowekamp2013design,
  title={The design of SimpleITK},
  author={Lowekamp, Bradley C and Chen, David T and Ib{\'a}{\~n}ez, Luis and Blezek, Daniel},
  journal={Frontiers in neuroinformatics},
  volume={7},
  pages={45},
  year={2013},
  publisher={Frontiers Media SA}
}

@InProceedings{DeTone_2018_SuperPoint,
author = {DeTone, Daniel and Malisiewicz, Tomasz and Rabinovich, Andrew},
title = {SuperPoint: Self-Supervised Interest Point Detection and Description},
booktitle = {Proceedings of the IEEE Conference on Computer Vision and Pattern Recognition (CVPR) Workshops},
year = {2018}
}

@InProceedings{Sarlin_2020_SuperGlue,
author = {Sarlin, Paul-Edouard and DeTone, Daniel and Malisiewicz, Tomasz and Rabinovich, Andrew},
title = {SuperGlue: Learning Feature Matching With Graph Neural Networks},
booktitle = {Proceedings of the IEEE/CVF Conference on Computer Vision and Pattern Recognition (CVPR)},
year = {2020}
}

@article{budelmann_histokatfusion_2022,
	series = {{ACROBAT} 2022},
	title = {{HISTOKATFUSION}. {IMAGE} {REGISTRATION} {FOR} {THE} {ACROBAT} {CHALLENGE}.},
	language = {en},
	author = {Budelmann, Daniel and Weiss, Nick and Heldmann, Stefan and Lotz, Johannes},
	year = {2022},
}

@article{wodzinski_aghsso_nodate,
	title = {{AGHSSO} - {Short} {Description} of the {Contribution} to the {ACROBAT} {Challenge}},
	language = {en},
	author = {Wodzinski, Marek and Jurgas, Artur and Marini, Niccolò and Atzori, Manfredo and Müller, Henning},
	year = {2022},
}

@article{gatenbee2023virtual,
  title={Virtual alignment of pathology image series for multi-gigapixel whole slide images},
  author={Gatenbee, Chandler D and Baker, Ann-Marie and Prabhakaran, Sandhya and Swinyard, Ottilie and Slebos, Robbert JC and Mandal, Gunjan and Mulholland, Eoghan and Andor, Noemi and Marusyk, Andriy and Leedham, Simon and others},
  journal={Nature communications},
  volume={14},
  number={1},
  pages={4502},
  year={2023},
  publisher={Nature Publishing Group UK London}
}

@InProceedings{leutenegger2011brisk,
  author={Leutenegger, Stefan and Chli, Margarita and Siegwart, Roland Y.},
  booktitle={2011 International Conference on Computer Vision}, 
  title={BRISK: Binary Robust invariant scalable keypoints}, 
  year={2011},
  pages={2548-2555},
  doi={10.1109/ICCV.2011.6126542}}

@inproceedings{raza2019deconvolving,
  title={Deconvolving convolutional neural network for cell detection},
  author={Raza, Shan E Ahmed and AbdulJabbar, Khalid and Jamal-Hanjani, Mariam and Veeriah, Selvaraju and Le Quesne, John and Swanton, Charles and Yuan, Yinyin},
  booktitle={2019 IEEE 16th International Symposium on Biomedical Imaging (ISBI 2019)},
  pages={891--894},
  year={2019},
  organization={IEEE}
}

@article{sirinukunwattana2016locality,
  title={Locality sensitive deep learning for detection and classification of nuclei in routine colon cancer histology images},
  author={Sirinukunwattana, Korsuk and Raza, Shan E Ahmed and Tsang, Yee-Wah and Snead, David RJ and Cree, Ian A and Rajpoot, Nasir M},
  journal={IEEE transactions on medical imaging},
  volume={35},
  number={5},
  pages={1196--1206},
  year={2016},
  publisher={IEEE}
}

@article{bentley1975multidimensional,
  title={Multidimensional binary search trees used for associative searching},
  author={Bentley, Jon Louis},
  journal={Communications of the ACM},
  volume={18},
  number={9},
  pages={509--517},
  year={1975},
  publisher={ACM New York, NY, USA}
}

@phdthesis{trahearn2017registration,
  title={Registration and multi-immunohistochemical analysis of whole slide images of serial tissue sections.},
  author={Trahearn, Nicholas},
  year={2017},
  school={University of Warwick}
}
\end{document}